\pdfoutput=1

\documentclass[11pt]{article}

\usepackage[]{acl}

\usepackage{times}
\usepackage{latexsym}

\usepackage[T1]{fontenc}

\usepackage[utf8]{inputenc}

\usepackage{microtype}
\usepackage{microtype}
\usepackage[utf8]{inputenc} 
\usepackage[T1]{fontenc}    
\usepackage{hyperref}       
\usepackage{url}            
\usepackage{booktabs,colortbl   }       
\usepackage{amsfonts}       
\usepackage{nicefrac}       
\usepackage{microtype}      
\usepackage{xcolor}         

\usepackage{balance}
\usepackage{bm}
\usepackage{amssymb}
\usepackage{url}
\usepackage{makecell}
\usepackage{colortbl}
\usepackage{textcomp}
\usepackage{CJK}
\usepackage{enumitem}
\usepackage{amsfonts}
\usepackage{caption}
\usepackage{float}
\usepackage{multirow}
\usepackage{color}
\usepackage{arydshln}
\usepackage{wrapfig}
\usepackage{graphicx}
\usepackage{amsmath}
\usepackage{enumitem}
\usepackage{graphics}
\usepackage{graphicx}
\usepackage{subfig}

\usepackage{url}            
\usepackage{booktabs}       
\usepackage{amsfonts}       
\usepackage{nicefrac}       
\usepackage{microtype}      
\usepackage{amsmath}
\usepackage{makecell}
\usepackage{caption}
\usepackage{float}
\usepackage{multirow}
\usepackage{color}
\usepackage{arydshln}
\usepackage[utf8]{inputenc} 
\usepackage[T1]{fontenc}    
\usepackage{nicefrac}       
\usepackage{microtype}      
\usepackage{url}
\usepackage{amsfonts}
\usepackage{amsmath}
\usepackage{amssymb}
\usepackage{amsthm}
\usepackage{mathrsfs}
\usepackage{tikz}
\usepackage{graphicx}
\usepackage{algpseudocode}
\usepackage{verbatim}
\usepackage{xcolor}
\usepackage{enumitem}
\usepackage{booktabs}
\usepackage{multirow}
\usepackage{todonotes}
\usepackage{footnote}
\usepackage{listings}
\usepackage[font=small]{caption}
\usepackage[linesnumbered,ruled]{algorithm2e}
\makesavenoteenv{tabular}
\definecolor{codeBackground}{rgb}{1.00,1.00,1.00}

\lstnewenvironment{python}[2]
{
  \lstset{
    title={#1},
    backgroundcolor=\color{codeBackground},
    numbers=left,
    stepnumber=1,
    basicstyle=\ttfamily\footnotesize,
    keywordstyle=\color{blue}\bfseries,
    showstringspaces=false,
    xleftmargin=10pt,
    language=python,
    frame=t,
  }
}
{}

\DeclareMathOperator*{\argmax}{arg\,max}

\title{Improving Neural Machine Translation by Denoising Training}

\author{
Liang Ding\\
The University of Sydney\\
\normalsize \texttt{ldin3097@sydney.edu.au}\\
\And
Keqin Peng\thanks{~~Work done when interning at JD Explore Academy.}\\
Beihang University\\
\normalsize \texttt{keqin.peng@buaa.edu.cn}\\
\And
Dacheng Tao\\
JD Explore Academy, JD.com\\
\normalsize \texttt{dacheng.tao@gmail.com}}

\begin{document}
\maketitle
\begin{abstract}
We present a simple and effective pretraining strategy \underline{D}en\underline{o}ising \underline{T}raining (DoT\footnote{DoT is one of the twin tricks for NMT (another is Bidirectional Training, namely BiT~\cite{Ding2021ImprovingNM}) that we proposed in IWSLT21 evaluation~\cite{Ding2021TheUS}.}) for neural machine translation. Specifically, we update the model parameters with source- and target-side denoising tasks at the early stage and then tune the model normally. Notably, our approach does not increase any parameters or training steps, requiring the parallel data merely. Experiments show that DoT consistently improves the neural machine translation performance across 12 \textit{bilingual} and 16 \textit{multilingual} directions (data size ranges from 80K to 20M). In addition, we show that DoT can complement existing data manipulation strategies, i.e. curriculum learning, knowledge distillation, data diversification, bidirectional training, and back-translation.
Encouragingly, we found that DoT outperforms costly pretrained model mBART~\cite{Liu2020MultilingualDP} in high-resource settings. Analyses show DoT is a novel in-domain cross-lingual pretraining
strategy, and could offer further improvements with task-relevant self-supervisions.
\end{abstract}

\section{Introduction}
Transformer~\cite{transformer} has been become the \textit{de facto} choice in neural machine translation (NMT)
due to its state-of-the-art performance~\cite{wmt19,wmt20,wmt21}.
However, an interesting study reveals that many Transformer modifications do not result in improved performance due to the lack of generalization~\cite{narang-etal-2021-transformer}. This finding is consistent with the recent call for data-centric AI in the ML community~\cite{ng6chat}, urging the NMT community pays more attention to how to effectively and efficiently exploit the supervisions from data rather than complicated modifications.

There has been a lot of works on NMT data manipulation to fully exploit the training data. \newcite{zhang2018empirical,platanios2019competence,liu2020norm,zhou-etal-2021-self,Ding2021Progressive} design difficulty metrics to enable the models to learn from easy to hard. 
\newcite{kim-rush-2016-sequence} propose sequence-level knowledge distillation for machine translation to acquire the refined knowledge from teachers.
\newcite{nguyen2019data} diversify the training data by using the predictions of multiple forward and backward models. 
Recently, \newcite{Ding2021ImprovingNM} initialize the translation system with a bidirectional system to obtain better performance.
However, they assume that the supervisions come from the correlation -- the basic properties of parallel data, between the source and target sentences, i.e. src$\leftrightarrow$tgt, ignoring the self-supervisions of the source or target sentences themselves.

In this work, we decide to find more self-supervisions from parallel data, which is hopefully complementary to existing data manipulation strategies.
Accordingly, we break the parallel data into two pieces of high-quality monolingual data, allowing us to design rich self-supervisions on both source and target side. We choose denoising as the self-supervision objective, i.e. \textit{denoising training} (\S\ref{sec:method}). The core idea is using a multilingual denoising system as the initialization for a translation system. 
Specifically, given the parallel language pair ``$\text{B}$: $\text{src}\rightarrow\text{tgt}$'', we can construct the denosing data ``$\text{M}_\mathrm{src}$: ${noised}(\mathrm{src})\rightarrow\mathrm{src}$'' and ``$\text{M}_\mathrm{tgt}$: ${noised}(\mathrm{tgt})\rightarrow\mathrm{tgt}$''. Then we update the parameters with denoising data $\text{M}_\mathrm{src}+\text{M}_\mathrm{tgt}$ in the early stage, and tune the model with parallel data $\text{B}$. 

We validated our approach on bilingual and multilingual benchmarks across different language families and sizes in \S\ref{subsec:results}. 
Experiments show DoT consistently improves the translation performance. We also show DoT can complement existing data manipulation strategies, i.e. back translation~\cite{caswell2019tagged}, curriculum learning~\cite{platanios2019competence}, knowledge distillation~\cite{kim-rush-2016-sequence}, data diversification~\cite{nguyen2019data} and bidirectional training~\cite{Ding2021ImprovingNM}.
Analyses in \S\ref{subsec:analysis} provide some insights about where the improvements come from: DoT is a simple in-domain cross-lingual pretraining strategy and can be enhanced with task-relevant self-supervisions.

\section{Denoising Training}
\label{sec:method}
\paragraph{Preliminary}
Given a source sentence $\bf x$, NMT models generate target word ${\bf y}_t$ conditioned on previously generated ${\bf y}_{<t}$, which can be formulated as:
\begin{equation}
    p({\bf y}|{\bf x})
    =\prod_{t=1}^{T}p({\bf y}_t|{\bf x},{\bf y}_{<t}; \theta)
    \label{eq:standard}
\end{equation}
where $T$ is the length of the target sequence and the parameters $\theta$ are trained to maximize the likelihood of training examples according to $\mathcal{L}(\theta) = \argmax_{\theta} \log p({\bf y}|{\bf x}; \theta)$. 
The training examples used to achieve the conditional estimation can be defined as
    $\text{B} = \{(\mathbf{x}_i, \mathbf{y}_i)\}^N_{i=1}$
, where $N$ is the total number of sentence pairs in the training data. 

\label{subsec:bidirectional}
\paragraph{Motivation}
The motivation of {\bf \emph{training with denoising data}} is when humans learn languages, one of the best practice for language acquisition is to correct the sentence errors~\cite{marcus1993negative}.
Motivated by it, \newcite{lewis2020bart} propose several noise functions and denoise them in end-to-end way. \newcite{Liu2020MultilingualDP} introduce this idea to the multilingual scenarios. Different from above monolingual pretraining approaches, we propose a simpler noise function and apply it to each side of the parallel data. 

\paragraph{Method}
We want the model to understand both the source- and target-side languages well before lexical translation and reordering~\cite{voita-etal-2021-language}.
For noise function ${noised}(\cdot)$, we apply the common noise-injection practices in Appendix~\ref{app:code}, i.e. removing, replacing, or nearby swapping one time for a random word with a uniform distribution in a sentence~\cite{edunov2018understanding}. Then size of the original parallel data doubled as follows:
\begin{align}
    \text{M}_{\mathrm{src}} &= \{{noised}(\mathbf{x}_i), \mathbf{x}_i\}^N_{i=1}\\
    \text{M}_{\mathrm{tgt}} &= \{{noised}(\mathbf{y}_i), \mathbf{y}_i\}^N_{i=1}
    \label{eq:denoise}
\end{align}
where $\text{M}_{\mathrm{src}}$ and $\text{M}_{\mathrm{tgt}}$ can be combined to update the end-to-end model. In doing so, $\theta$ in Eq.~\ref{eq:standard} can be updated by denoising both the source and target data, then the denoising objective becomes:
\begin{align}
    {\mathcal{L}_\text{DoT}}(\theta) &= \overbrace{\argmax_{\theta} \log p({\bf x}|{{noised}({\bf x})}; \theta)}^{\text{Source Denoising}: {\mathcal{L}_{\theta}^{S}}}\\
    &+ \underbrace{\argmax_{\theta} \log p({\bf y}|{{noised}({\bf y})}; \theta)}_{\text{Target Denoising}: {\mathcal{L}_{\theta}^{T}}}
    \label{eq:dptloss}
\end{align}
where the source denoising objective ${\mathcal{L}_{\theta}^{S}}$ and target donising objective ${\mathcal{L}_{\theta}^{T}}$ are optimized iteratively. The pretraining can store knowledge of the source and target languages into the shared model parameters, which may help {\em better} and {\em faster} learning further tasks. 
Following \citet{Ding2021ImprovingNM}, we early stop denoising training at 1/3 of the total steps, and tune the model normally with the rest of 2/3 training steps. This process can be formally denoted as such pipeline: $\text{M}_{\mathrm{src}}+\text{M}_{\mathrm{tgt}}\rightarrow{\text{B}}$.
There are many possible ways to implement the general idea of denoising training. The aim of this paper is not to explore the whole space but simply to show that one fairly straightforward implementation works well and the idea is reasonable.

\begin{table*}[t]
    \centering
    \setlength{\tabcolsep}{3.2pt}
    \scalebox{0.9}{
    \begin{tabular}{lllllllllllll}
    \toprule
    \textbf{Data Source} &
    \multicolumn{2}{c}{\textbf{IWSLT14}} &
    \multicolumn{2}{c}{\textbf{WMT16}}  &
    \multicolumn{2}{c}{\textbf{IWSLT21}} & 
    \multicolumn{2}{c}{\textbf{WMT14}} & \multicolumn{2}{c}{\textbf{WMT20}} &
    \multicolumn{2}{c}{\textbf{WMT17}} \\
    \textbf{Size} &
    \multicolumn{2}{c}{160K} &
    \multicolumn{2}{c}{0.6M} &
    \multicolumn{2}{c}{2.4M} &
    \multicolumn{2}{c}{4.5M} &
    \multicolumn{2}{c}{13M} &
    \multicolumn{2}{c}{20M}\\
    \cdashline{2-13}
    \textbf{Direction} & \bf En-De &\bf De-En & \bf En-Ro & \bf Ro-En  & \bf En-Sw & \bf Sw-En & \bf En-De &\bf De-En & \bf Ja-En &\bf En-Ja & \bf Zh-En &\bf En-Zh\\
    \midrule
    {\bf Transformer} & 29.2 & 35.1  & 33.9& 34.1  & 28.8 & 48.5 & 28.6 & 32.1 & 20.4 & 18.2 & 23.7 & 33.2\\
    {\bf ~~~~+DoT} & 29.8$^\dagger$ & 36.1$^\ddagger$ & 35.0$^\ddagger$ & 35.5$^\ddagger$ & 29.3$^\ddagger$ & 49.6$^\ddagger$ & 29.5$^\ddagger$ & 32.7$^\dagger$ & 20.9$^\dagger$ &19.1$^\ddagger$ &24.7$^\ddagger$ &33.6\\
    \bottomrule
    \end{tabular}}
    \caption{Performance on several widely-used bilingual benchmarks, including IWSLT14 En$\leftrightarrow$De, WMT16 En$\leftrightarrow$Ro, IWSLT21 En$\leftrightarrow$Sw, WMT14 En$\leftrightarrow$De, WMT20 Ja$\leftrightarrow$En and WMT17 Zh$\leftrightarrow$En. Among them, Ja-En and Zh-En are distant language pairs. ``$^{\ddagger/\dagger}$'' indicates significant difference ($p < 0.01/ 0.05$) from corresponding baselines.}
    \label{tab:main-results}
\end{table*}


\section{Experiments}
\subsection{Setup}
\label{subsec:setup}
\paragraph{Bilingual Data}
Main experiments in Tab.~\ref{tab:main-results} are conducted on 6 translation datasets: IWSLT14 English$\leftrightarrow$German~\cite{nguyen2019data}, WMT16 English$\leftrightarrow$Romanian~\cite{gu2018non}, IWSLT21 English$\leftrightarrow$Swahili\footnote{\url{https://iwslt.org/2021/low-resource}}, WMT14 English$\leftrightarrow$German~\cite{transformer}, WMT20 Japanese$\leftrightarrow$English\footnote{\url{http://www.statmt.org/wmt20}} and WMT17 Chinese$\leftrightarrow$English~\cite{hassan2018achieving}. The data sizes can be found in Tab.~\ref{tab:main-results}, ranging from 160K to 20M. Notably, Japanese$\leftrightarrow$English and Chinese$\leftrightarrow$English are two distant and high-resource language pairs. The monolingual data used for back translation in Tab.~\ref{tab:complementary} is randomly sampled from publicly available News Crawl corpus\footnote{\url{http://data.statmt.org/news-crawl/}}.
We use same valid\& test sets with previous works for fair comparison excepts IWSLT21 English$\leftrightarrow$Swahili, where we sample 5K/ 5K sentences from the training set as valid/ test sets.
We preprocess all data besides Japanese$\leftrightarrow$English via BPE~\cite{Sennrich:BPE} with 32K merge operations. For Japanese$\leftrightarrow$English, we filter the parallel data with Bicleaner~\cite{sanchez-cartagena-etal-2018-prompsits} and apply the SentencePiece~\cite{kudo-richardson-2018-sentencepiece} to generate 32K subwords. 
\begin{table}[t]
    \centering
    \scalebox{1.0}{
    \begin{tabular}{lcccc}
    \toprule
    \bf Languages&\bf Fa &\bf Pl &\bf Ar &\bf He \\
    \bf Size &89K &128K &140K &144K\\
    \midrule
    \bf Transformer & 17.1 & 16.4 & 21.3 & 28.8 \\   
    \bf ~~~~+DoT    & 18.2$^\ddagger$ & 17.5$^\dagger$ & 22.8$^\ddagger$ & 30.7$^\ddagger$ \\
    \midrule 
    \midrule
    \bf Lang.&\bf Nl &\bf De &\bf It &\bf Es \\
    \bf Size &153K &160K &167K &169K\\
    \midrule
    \bf Transformer & 31.5 & 28.5 & 29.3 & 34.9 \\
    \bf ~~~~+DoT    & 33.3$^\ddagger$ & 29.6$^\ddagger$ & 31.2$^\ddagger$ & 36.5$^\ddagger$ \\
    \bottomrule
    \end{tabular}}
    \caption{Performance on IWSLT multilingual task. For simplicity, we report the average BLEU of En$\rightarrow$X and X$\rightarrow$En within one language. For significance, we compare the translation concatenation of En$\rightarrow$X and X$\rightarrow$En and corresponding concatenated references.}
    \label{tab:multilingual}
\end{table}
\paragraph{Multilingual Data}
We follow \citet{lin2021learning} to collect eight English-centric multilingual language pairs from IWSLT14\footnote{\url{https://wit3.fbk.eu/}}, including Farsi (Fa), Polish (Pl), Arabic (Ar), Hebrew (He), Dutch (Nl), German (De), Italian (It), and Spanish (Es).
Following~\citet{tan2019multilingual}, we apply BPE with 30K merge operations, and use over-sampling strategy to balance the training data distribution with temperature of $T=2$. 
The hyper-parameters of \textit{removing} and \textit{replacing} are set as ratio$=$0.1, and the \textit{nearby swapping} as span$=$3 due to their better performance in our preliminary studies. For evaluation, we use tokenized BLEU~\cite{papineni2002bleu} as the metric for bilingual tasks excepts English$\rightarrow$Chinese and Japanese$\leftrightarrow$English, where we report SacreBLEU~\cite{post-2018-call}. The \textit{sign-test}~\cite{collins2005clause} was utilized for statistical significance test.

\paragraph{Model}
We validated our proposed DoT on Transformer\footnote{\url{https://github.com/pytorch/fairseq}}~\cite{transformer}. All bilingual tasks are trained on Transformer-\textsc{Big} except IWSLT14 En$\leftrightarrow$De and WMT16 En$\leftrightarrow$Ro (trained on Transformer-\textsc{Base}) because of their extremely small data size. For multilingual experiments, we closely follow previous work~\cite{wu2019pay} to adopt smaller\footnote{Transformer-Base with $d_{ff}=1024$ and $n_{head}=4$} Transformer-\textsc{Base} due to the small-scale data volume of the IWSLT multilingual dataset. For fair comparison, we set beam size and length penalty as 5 and 1.0 for all language pairs.
Notably, our data-level approaches neither modify model structure nor add extra FLOPS, thus they are feasible to deploy on any frameworks (e.g. DynamicConv~\cite{wu2019pay} and non-autoregressive translation~\cite{gu2018non,Ding2020Rejuvenating}) and other sequence-to-sequence tasks (e.g. grammar error correction and summarization~\cite{Liu2021UnderstandingAI}). We will explore them in the future works. 

Someone may doubt that why not introduce more powerful pretrained language models, e.g. mBART~\cite{Liu2020MultilingualDP}, as the baseline. We politely argue that it is not fair to directly compare DoT with mBART in the main results because: 1) Our DoT consumes significantly less parameters compared to mBART, 49M\textasciitilde200M vs. 610M; and 2) Our DoT uses only the parallel data, while mBART uses TB-level text during pretraing.
In addition, to show the effectiveness and efficiency of our DoT, we report the comparison results between mBART and DoT in \S\ref{subsec:analysis} (see Tab.~\ref{tab:mbart}). We show that although mBART performs well on low-resource settings, however, it achieve worse performance on the high-resource settings.

\paragraph{Training}
For Transformer-\textsc{Big} models, we adopt large batch strategy~\cite{edunov2018understanding} (i.e. 458K tokens/batch) to optimize the performance. The learning rate warms up to $1\times10^{-7}$ for 10K steps, and then decays for 30K (data volumes range from 2M to 10M) / 50K (data volumes large than 10M) steps with the cosine schedule; For Transformer-\textsc{Base}, we empirically adopt 65K tokens/batch for small data sizes, e.g. IWSLT14 En$\rightarrow$De and WMT16 En$\rightarrow$Ro. The learning rate warms up to $1\times10^{-7}$ for 4K steps, and then decays for 26K steps.
For regularization, we tune the dropout rate from [0.1, 0.2, 0.3] based on validation performance, and apply weight decay with 0.01 and label smoothing with 0.1. We use Adam optimizer ~\citep{kingma2015adam} to train models. We evaluate the performance on an ensemble of last 10 checkpoints to avoid stochasticity. All models were trained on \texttt{NVIDIA DGX A100} cluster.

Someone may doubt that DoT heavily depends on how to properly set the early-stop steps. To dispel the doubt, we investigate whether our approach is robust to different early-stop steps. In preliminary experiments, we tried several simple fixed early-stop steps according to the size of training data (e.g. training 40K En-De and early stop at 10K/ 15K/ 20K, respectively). We found that both strategies achieve similar performances. Thus, we decide to choose a simple threshold (i.e. 1/3 of the total training steps) for better reproduction.
\begin{table*}[t]
    \centering
    \scalebox{1.0}{
    \begin{tabular}{lll}
    \toprule
    \bf Types &\bf Model & \bf BLEU \\
    \midrule
    &\textbf{Transformer-\textsc{Big}}/~+DoT & 28.6/ 29.5$^\ddagger$\\
    \hdashline
    \multirow{4}*{\bf Parallel}&\textsc{~~+Curriculum Learning}~\cite{platanios2019competence}/~+DoT & 29.4/ 29.8$^\dagger$\\
    &\textsc{~~+Knowledge Distillation}~\cite{kim-rush-2016-sequence}/~+DoT & 29.3/ 29.7\\
    &\textsc{~~+Data Diversification}~\cite{nguyen2019data}/~+DoT & 30.1/ 30.6$^\dagger$\\
    &\textsc{~~+Bidirectional Training}~\cite{Ding2021ImprovingNM}/~+DoT & 29.7/ 29.9\\
    \hdashline
    \multirow{1}*{\bf Monolingual} &\textsc{~~+Back Translation}~\cite{caswell2019tagged}/~+DoT & 30.5/ 31.1$^\dagger$\\
    \bottomrule
    \end{tabular}}
\caption{Complementary to other works. ``/+DoT'' means combining DoT with corresponding data manipulation works, and BLEU scores of DoT followed their counterparts with ``/''. Experiments are conducted on WMT14 En-De.}
\label{tab:complementary}
\end{table*}

\begin{table}[tb]
    \centering
    \scalebox{1}{
    \begin{tabular}{llllll}
        \toprule
         \textbf{Scales} &\rm 100K &\rm 1M &\rm 5M &\rm 10M &\rm 20M  \\
         \midrule
         \bf Random &10.8 &16.4 &21.5 &27.3 &33.2\\
         \midrule
         \bf mBART  &\cellcolor{blue!62}17.4 &\cellcolor{blue!41}20.5 &\cellcolor{blue!28}22.3 &\cellcolor{red!17}26.8 &\cellcolor{red!58}31.4\\
         \bf DoT    &\cellcolor{blue!37}12.9 &\cellcolor{blue!30}18.1 &\cellcolor{blue!26}22.1 &\cellcolor{blue!18}27.8 &\cellcolor{blue!16}33.6\\
        \bottomrule
    \end{tabular}}
    \caption{Translation performance comparison between mBART-based models and our proposed DoT across different data scales. \textcolor{blue}{Blue} color represents improved performance over Random, while \textcolor{red}{Red} means reduction. Shades of cell color mean the significance degree.}
    \label{tab:mbart}
\end{table}
\subsection{Results}
\label{subsec:results}
\paragraph{Results on Different Data Scales}
We experimented on 12 language directions, including IWSLT14 En$\leftrightarrow$De, WMT16 En$\leftrightarrow$Ro, IWSLT21 En$\leftrightarrow$Sw, WMT14 En$\leftrightarrow$De, WMT20 Ja$\leftrightarrow$En and WMT17 Zh$\leftrightarrow$En. Tab.~\ref{tab:main-results} reports the experimental results, DoT achieves significant improvements over strong baseline Transformer in 8 out of 12 directions under the significance test $p<0.01$, and the other 3 directions also show promising performance under the significance test $p<0.05$, showing the effectiveness of our method across data scales.

\paragraph{Results on Distant Language Pairs}
We report the results of DoT on Ja$\leftrightarrow$En and Zh$\leftrightarrow$En language pairs, which belong to different language families (i.e. Japonic, Indo-European and Sino-Tibetan). As shown in Tab.~\ref{tab:main-results}, DoT significantly improves the translation quality in all cases. In particular, DoT achieves averaged +0.7 BLEU points improvement over the baselines, showing the effectiveness and universality of our method across language pairs.

\paragraph{Results on Multilingual Translation Tasks}
Cross-lingual pretrained models are expected to learn better representations when given more languages~\cite{Liu2020MultilingualDP}. To verify this hypothesis, we report the multilingual results in Tab.~\ref{tab:multilingual}. As seen, DoT consistently and significantly outperforms the strong baseline on all language pairs, confirming the effectiveness of DoT under multilingual MT scenario.

\paragraph{Complementary to Related Work}
To illustrate the complementarity between DoT and related data manipulation works, we list one \textbf{monolingual} data-based approach: {Tagged Back Translation} (\textbf{BT},~\citealt{caswell-etal-2019-tagged}) combines the synthetic data generated with target-side \textit{monolingual data} and parallel data, and four representative \textbf{parallel} data-based approaches: 
a) {Competence based Curriculum Learning} (\textbf{CL},~\citealt{platanios2019competence}) trains samples with easy to hard order, where the difficulty metric is the competence;
b) {Knowledge Distillation} (\textbf{KD},~\citealt{kim-rush-2016-sequence}) trains the model with sequence-level distilled \textit{parallel data};
c) {Data Diversification} (\textbf{DD},~\citealt{nguyen2019data}) 
diversifies the data by applying KD and BT on \textit{parallel data}; 
d) {Bidirectional Training} (\textbf{BiT},~\citealt{Ding2021ImprovingNM}) pretrains the model with bidirectional \textit{parallel data}.
As seen in Tab.~\ref{tab:complementary}, DoT is a plug-and-play strategy to yield further improvements. 

Here we only provide the empirical evidence for the complementary between our pretraining approach and conventional data manipulation strategies, more insights like \citet{Liu2021OnTC} investigated should be explored in the future.

\subsection{Analysis}
\label{subsec:analysis}
We conducted analyses on WMT17 En$\Rightarrow$Zh dataset to better understand where the improvement comes from.
\paragraph{DoT works as an in-domain pretrain strategy}
\newcite{song2019mass,Liu2020MultilingualDP} reconstruct the abundant unlabeled data with sequence-to-sequence language models, offering the benefit for the low-resource translation tasks~\cite{cheng2021self}. However, such benefits are hard to enjoy in high-resource settings in part due to the catastrophic forgetting~\cite{french1999catastrophic} and domain discrepancy between pretrain and finetune~\cite{gururangan-etal-2020-dont,anonymous2021arr}. Actually, DoT can be viewed as a simple in-domain pretraining strategy, where the pretraining corpora are exactly matched with the downstream machine translation task. We randomly sample different data scales to carefully compare the effects of pretrained mBART \footnote{\url{https://dl.fbaipublicfiles.com/fairseq/models/mbart/mbart.cc25.v2.tar.gz}}
\cite{Liu2020MultilingualDP} and our proposed DoT. The mBART is used to initialize the downstream bilingual models with the officially given scripts\footnote{\url{https://github.com/pytorch/fairseq/blob/main/examples/mbart/README.md}}. To maximize the effect of mBART, we grid-searched the ``\texttt{--update-freq}''\footnote{Increasing the batch size  during finetuning mBART for large scale datasets is essential to achieve better performance.} from 2 to 16 to simulate different batch sizes. As for our DoT, we employ Transformer-\textsc{Base} for 100K and 1M data scales, and Transformer-\textsc{Big} for other larger datasets, i.e. 5M, 10M, and 20M. notably, under such experimental settings, our DoT consumes \textasciitilde200M parameters for the high resource setting, while the mBART contains significantly more parameters, i.e. \textasciitilde610M, for all settings. 

Tab.~\ref{tab:mbart} shows that although mBART achieves significant higher performance under extremely low-resource setting, e.g. 100K, it undermines the model performance under high-resource settings, e.g. $>=$10M, while our DoT strategy consistently outperforms ``Random'' setting. We attribute the better performance on resource-rich settings to the ``in-domain pretrain'', which is consistent with the findings of \citet{gururangan-etal-2020-dont,anonymous2021arr}. 


\paragraph{DoT works better with task-relevant signals}
Recall that we simply employ several straightforward noises, e.g. word drop, word mask and span shuffle, during sequence-to-sequence reconstruction, which are then serve as the self-supervision signals. \newcite{gururangan-etal-2020-dont} show that task-relevant signals are important during finetuning the pretrained models. In machine translation, reconstructing the cross-lingual knowledge is intuitively more informative than randomly drop, mask, and shuffling. We therefore deliberately design a translation-relevant denoising signal to provide more insight for our DoT. Concretely, during word masking, we input the code-switched words rather than [MASK] token for the input, where we closely follow \citet{yang2020csp} to achieve the code-switching. We found that task-relevant signals enable the model to achieve further improvement by 0.3 BLEU points, confirming our claim.

\section{Conclusion}
In this paper, we present a simple self-supervised pretraining approach for neural machine translation with parallel data merely.
Extensive experiments on bilingual and multilingual language pairs show that our approach DoT consistently and significantly improves translation performance, and can complement well with existing data manipulation methods. In-depth analyses indicate that DoT works as an in-domain pretrain strategy, and can be a better alternative to costly large-scale pretrained models, e.g. mBART.

We hope our proposed twin tricks for NMT, i.e. DoT and BiT~\cite{Ding2021ImprovingNM}, could facilitate the MT researchers who aim to develop a system under constrained data resources.

\section*{Acknowledgements}
We would thank anonymous reviewers of ARR Nov. for their considerate proofreading and valuable comments.

\bibliography{anthology,custom}
\bibliographystyle{acl_natbib}

\appendix
\newpage
\onecolumn{
\section{Appendix}
\subsection{\label{app:code}Code for Noise Function in PyTorch}
\begin{python}{\texttt{Noises include removing, replacing and nearby swapping}}
import argparse
import fileinput
import random
import torch

def main():
    parser = argparse.ArgumentParser(\
        description='Command-line script to add noise to data')
    parser.add_argument('-wd', help='Word dropout', default=0.1, type=float)
    parser.add_argument('-wb', help='Word blank', default=0.1, type=float)
    parser.add_argument('-sk', help='Shufle k words', default=3, type=int)
    args = parser.parse_args()
    wd = args.wd
    wb = args.wb
    sk = args.sk

    for s in fileinput.input('-'):
        s = s.strip().split()
        if len(s) > 0:
            s = word_shuffle(s, sk)
            s = word_dropout(s, wd)
            s = word_blank(s, wb)
        print(' '.join(s))

def word_shuffle(s, sk):
    noise = torch.rand(len(s)).mul_(sk)
    perm = torch.arange(len(s)).float().add_(noise).sort()[1]
    return [s[i] for i in perm]

def word_dropout(s, wd):
    keep = torch.rand(len(s))
    res = [si for i,si in enumerate(s) if keep[i] > wd ]
    if len(res) == 0:
        return [s[random.randint(0, len(s)-1)]]
    return res

def word_blank(s, wb):
    keep = torch.rand(len(s))
    return [si if keep[i] > wb else 'MASK' for i,si in enumerate(s)]

if __name__ == '__main__':
    main()
\end{python}
}

\end{document}